# HIERARCHICAL ATTENTION MODEL FOR IMPROVED MACHINE COMPREHENSION OF SPOKEN CONTENT

*Wei Fang, Jui-Yang Hsu, Hung-yi Lee, Lin-Shan Lee*

Department of Electrical Engineering
National Taiwan University

b02901054@ntu.edu.tw, b02901085@ntu.edu.tw, hungyilee@ntu.edu.tw, lslee@gate.sinica.edu.tw

## ABSTRACT

Multimedia or spoken content presents more attractive information than plain text content, but the former is more difficult to display on a screen and be selected by a user. As a result, accessing large collections of the former is much more difficult and time-consuming than the latter for humans. It's therefore highly attractive to develop machines which can automatically understand spoken content and summarize the key information for humans to browse over. In this endeavor, a new task of machine comprehension of spoken content was proposed recently. The initial goal was defined as the listening comprehension test of TOEFL, a challenging academic English examination for English learners whose native languages are not English. An Attention-based Multi-hop Recurrent Neural Network (AMRNN) architecture was also proposed for this task, which considered only the sequential relationship within the speech utterances. In this paper, we propose a new Hierarchical Attention Model (HAM), which constructs multi-hopped attention mechanism over tree-structured rather than sequential representations for the utterances. Improved comprehension performance robust with respect to ASR errors were obtained.

*Index Terms*— spoken question answering, TOEFL, deep learning, attention model

## 1. INTRODUCTION

With the popularity of shared videos, social networks, online courses, etc., the quantity of multimedia or spoken content is growing much faster beyond what human beings can view or listen to. Accessing large collections of multimedia or spoken content is difficult and time-consuming for humans, even if these materials are more attractive for humans than plain text information. Hence, it is desirable that machines can automatically listen to and understand the spoken content, and extract or even visualize the key information for humans. An initial attempt towards the above goal of machine comprehension of spoken content was presented recently in an initial task [1],

```
Story
......The idea of primary colors does not exist until
about 2 years ago. Then the dominant theory about color
is one that has been proposed by Isaac Newton......But
he made no mention of primary colors......
(audio story)
```
```
Question
Why does the professor mention Issac Newton?
```
```
Choices
A. To show the similarity between early ideas in art
   and early ideas in science
B. To explain why mixing primary colors does not
   produce satisfactory secondary colors
C. To provide background information for the theory of
   primary colors
D. To point out the first person to propose a theory of
   primary colors
```

**Fig. 1**: An example problem set of TOEFL listening comprehension test. The story is given in audio (manual transcription shown). The question and choices are in text.

in which the machine is given an audio story, and required to answer the questions related to that audio story, as an evaluation regarding how well the machine comprehends the audio story. TOEFL listening comprehension test is such a task but for human English learners whose native languages are not English. An Attention-based Multi-hop Recurrent Neural Network (AMRNN) framework for the above TOEFL task was proposed, and encouraging initial results were reported recently [1]. This paper presents an improved framework with better results over the same task.

The listening comprehension task considered here is highly related to Spoken Question Answering (SQA) [2, 3]. In SQA, when the users enter questions in either text or spoken form, the machine needs to find the answer from some audio files. SQA usually worked with ASR transcripts of the spoken content, and used information retrieval (IR) techniques [4] or relied on knowledge bases [5] to find the proper answer. Sibyl [6] is a factoid SQA system, while Question Answering in Speech Transcripts (QAST) [7] has been a well-known evaluation program for years. However, most previous works on SQA mainly focused on factoid question answering like "What is the name of the highest mountain in Taiwan?" Sometimes this kind of questions may be correctly answered by simply extracting the key terms from a properly chosen utterance without understanding the given spoken content. More difficult questions that cannot be answered without understanding the whole spoken content seemed rarely dealt with previously. On the other hand, most Question Answering works focused on understanding text documents [8–11]. Even though MovieQA [12] tried to answer questions related to movies, they only used the text and images in the movies. Machine comprehension of spoken content thus remains to be a less investigated problem.

On the other hand, neural models have been extensively explored for question answering tasks [13–16]. Specifically, reasoning systems incorporating memory and attention mechanisms such as the Memory Network (MemNN) [17] were shown to be very successful, and the End-to-end Memory Network (MemN2N) [18], a variant of MemNN, can be trained end-to-end without labeled supporting facts. In these models, the sentences were encoded by bag-of-words (BoW) representations, without considering the word order. The Attention-based Multi-hop Recurrent Neural Network (AMRNN) mentioned above in the previous work [1] utilizes the attention mechanism with recurrent neural networks (RNN) [19] to construct sentence representations considering the word order, but didn't take the syntactic structure of sentences into account yet. Recently, tree-structured models [20] obtained from the syntactic structures of the sentences were shown to be able to produce more robust representations and capture better semantics in certain tasks. But they have not been applied on question answering tasks with attention mechanism.

## 2. TASK DESCRIPTION AND CONTRIBUTIONS

In this paper, we take the TOEFL listening comprehension test as the corpus for experiments [1]. TOEFL tests knowledge and skills of academic English for global English learners whose native languages are not English. Each example problem set consists of an audio story and a question with four answer choices. Among these choices, one or two of them are correct. An example is shown in Fig. 1. The machine has to select the correct answer(s) out of the four choices. The questions here are not very easy because the answer cannot be found by simply matching the question and the choices without understanding the story. For example, there are questions regarding the gist of the story or the conclusion for the conversation. So this is a relatively challenging task for state-of-the-art spoken language understanding technologies, and the proposed approaches should be general enough to tackle different question types.

In this paper, we propose a Hierarchical Attention Model (HAM) to construct tree-structured sentence representations for sentences from their parsing trees and estimate attention weights on different nodes of the hierarchies. This model is evaluated on the above TOEFL listening comprehension test. The experiments showed improved performance over existing baselines including the previous work using AMRNN [1].

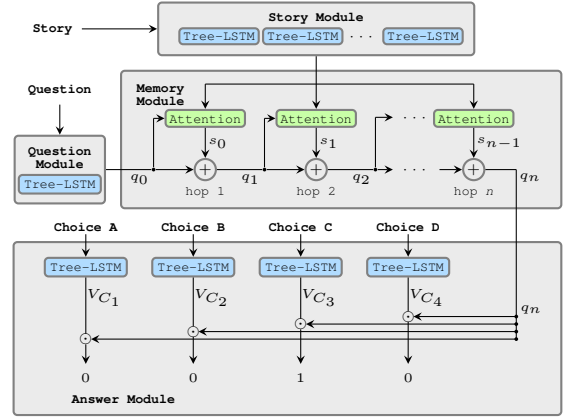

**Fig. 2**: The proposed Hierarchical Attention Model (HAM). The Tree-LSTM network architecture is shown in Fig. 3, and the attention mechanism is shown in Fig. 4.

## 3. HIERARCHICAL ATTENTION MODEL (HAM)

The proposed Hierarchical Attention Model (HAM) is shown in Fig. 2 in the form matched to the TOEFL task. In this model, tree-structured long short-term memory networks (Tree-LSTM, small blue blocks in Fig. 2) is used to obtain the representations for the sentences and phrases in the audio stories, questions and choices. The detailed operation of a Tree-LSTM is shown in Fig. 3 and explained below. The story module on the top of Fig. 2 computes tree-structured representations for the sentences in the transcriptions of the input audio story using Tree-LSTMs. The question module on the middle left generates a question vector representation from the word sequence of the question. The memory module in the middle includes attention modules (small green blocks) to draw question-related attention weights for different nodes of the sentences in the story. The detailed operation of the attention module is in Fig. 4 and explained below. Finally, the answer module at the bottom evaluates the confidence scores of the answer choices and generates the answer. The details of the different components of the model are given below.

### 3.1. Tree-LSTM

Two variants of Tree-LSTM can be used: the *Child-Sum Tree-LSTM* and the *N-ary Tree-LSTM*. A Child-Sum Tree-LSTM over a dependency tree is referred to as a Dependency Tree-LSTM, which is used here due to its relatively compact structure. The dependency tree structure of a sentence "The conventions may vary" is shown on the left part of Fig. 3, in which each node corresponds to a word (head word of the node) in the sentence.

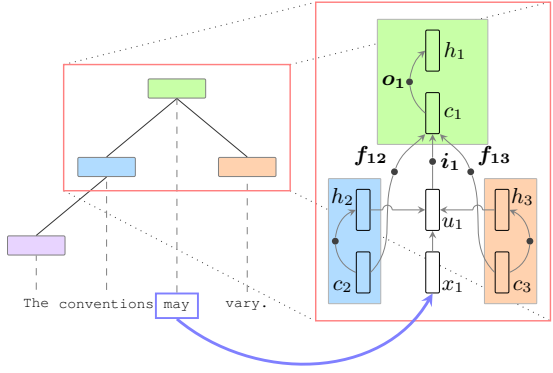

**Fig. 3**: Computation of the memory cell $c_1$ and hidden state $h_1$ of a Tree-LSTM node ($j = 1$) with two children ($k = 2, 3$).

Tree-LSTM generates a vector representation for each node in the dependency tree[1] based on the vector representations of its child nodes as in the right part of Fig. 3. Each node $j$ has a hidden state $h_j$ as its vector representation and a set of memory cells $c_j$. Just as the standard LSTM [22], it has the input gates $i_j$ and output gates $o_j$ for the memory cells, and the forget gates $f_{jk}$ that controls the information flowing in from its child node $k$. It also takes an input $x_j$, which is the vector representation of the head word of the node.

The hidden state $h_j$ at node $j$ is the representation for a phrase consisting of words in the subtree rooted at node $j$. Below is how $h_j$ of a node $j$ is obtained from its child nodes $k$. First, the hidden states of the child nodes are summed,

$$\tilde{h}_j = \sum_{k \in C(j)} h_k, \quad (1)$$

where $C(j)$ denotes the set of children of node $j$. The representation of the head word $x_j$ of node $j$ and $\tilde{h}_j$ in (1) are used to control the input gates $i_j$ and output gates $o_j$,

$$i_j = \sigma\big(W^{(i)} x_j + U^{(i)} \tilde{h}_j + b^{(i)}\big), \quad (2)$$
$$o_j = \sigma\big(W^{(o)} x_j + U^{(o)} \tilde{h}_j + b^{(o)}\big). \quad (3)$$

Each child $k$ in $C(j)$ has its own forget gate $f_{jk}$ obtained from $h_k$ and $x_j$,

$$f_{jk} = \sigma\big(W^{(f)} x_j + U^{(f)} h_k + b^{(f)}\big). \quad (4)$$

Finally, the hidden state $h_j$ and memory cells $c_j$ are obtained as the following:

$$u_j = \tanh\big(W^{(u)} x_j + U^{(u)} \tilde{h}_j + b^{(u)}\big), \quad (5)$$
$$c_j = i_j * u_j + \sum_{k \in C(j)} f_{jk} * c_k, \quad (6)$$
$$h_j = o_j * \tanh(c_j), \quad (7)$$

where $*$ denotes elementwise multiplication.

[1]Dependency parses produced by the Stanford Neural Network Dependency Parser [21].

### 3.2. Story and Question Modules

The story module and the question module produce representations for the sentences respectively in the story and in the question with the Tree-LSTM as explained in the subsection 3.1. For story module, the hidden vectors for all nodes in the tree structures of each sentence in the story are stored for future use. On the other hand, the question module produces the hidden states of the root nodes, $V_{S_i}$, of the Tree-LSTMs for the sentences $S_i$ in the question [2]. The question vector $V_Q$ is the sum of $V_{S_i}$ for all $S_i$ in a question, to be used below.

### 3.3. Memory Module

The memory module aims to extract the information in the story relevant to the question $V_Q$ based on representations obtained from the story module. It consists of two components: the attention mechanism and the multi-hopping.

**Attention Mechanism**

The attention mechanism is shown in Fig. 4. Let $O = \{o_1, o_2, ..., o_T\}$ be the set of vector representations for the story. There are two different ways to obtain these vectors:

- **Phrase-level**: $O = \{o_1, o_2, ..., o_T\}$, where each $o_t$ is the hidden state of a node in the Tree-LSTM of the sentences, or each $o_t$ represents a phrase. So $T$ is much larger than the number of sentences in the story. This is shown in the lower part of Fig. 4.

- **Sentence-level**: $O = \{o_1, o_2, ..., o_T\}$, where each $o_t$ is the hidden state of the root node of the Tree-LSTM over a sentence in the story, or each $o_t$ represents a sentence. So $T$ is equal to the number of sentences.

The vectors in the set $O$ are first transformed into memory vectors $M = \{m_1, m_2, ..., m_T\}$ and evidence vectors $C = \{c_1, c_2, ..., c_T\}$ by the embedding matrices $W^{(m)}$ and $W^{(c)}$,

$$m_t = W^{(m)} o_t, \quad c_t = W^{(c)} o_t. \quad (8)$$

The question vector $V_Q$ obtained in the question module is also transformed into an initial query vector $q_0$ by an embedding matrix $W^{(q)}$.

$$q_0 = W^{(q)} V_Q. \quad (9)$$

Cosine similarity is used to compute the attention score $\eta_t$ between the query vector $q_0$ and each memory vector $m_t$, which is further normalized by a *SoftMax* function, as shown in the upper left part of Fig. 4, to give the attention weights $\alpha = (\alpha_1, \alpha_2, ..., \alpha_T)$,

$$\eta_t = q_0 \odot m_t, \quad \alpha_t = \frac{e^{\eta_t}}{\sum_{i=1}^{T} e^{\eta_i}}, \quad (10)$$

[2]A question may have multiple sentences.

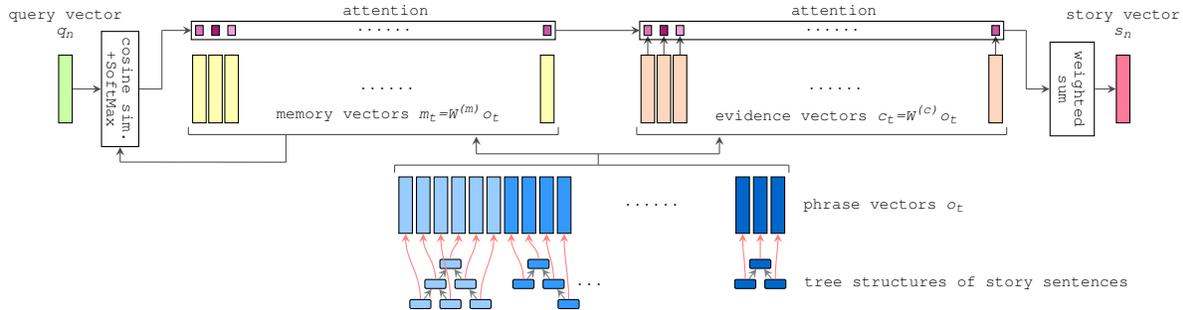

**Fig. 4**: Attention mechanism of the memory module. Phrase-level attention is shown in the figure.

where $\odot$ denotes cosine similarity. Each attention weight $\alpha_t$ corresponds to a memory vector $m_t$ and an evidence vector $c_t$ that represent a phrase or a sentence. The story vector $s_0$ is then the weighted sum of the evidence vectors $c_t$ with the attention as weights, as in the upper right part of Fig. 4,

$$s_0 = \sum_{t=1}^{T} \alpha_t c_t. \qquad (11)$$

**Multi-hopping**

As shown in Fig. 2, the sum of the initial query vector $q_0$ and the story vector $s_0$ obtained in (11) can be used as a new query vector $q_1$ to compute a new set of attention weights in exactly same way, and subsequently a new story vector $s_1$ can be obtained. This process can be repeated, allowing better focusing in the key information. The output of the final hop $q_n$ is the memory module output.

### 3.4. Answer Module

Suppose there are $K$ answer choices with $N$ correct answers. The answer module encodes the answer choices into choice vectors $V_{C_i}$. Cosine similarities between the memory module output $q_n$ and the choice vectors $V_{C_i}$, after passing through a SoftMax layer, give the predicted choice distribution $\hat{p}$. The target distribution $p$ is defined as:

$$p_i = \begin{cases} \frac{1}{N}, & \text{if choice } i \text{ is a correct answer} \\ 0, & \text{otherwise} \end{cases} \qquad (12)$$

for $1 \leq i \leq K$. The KL-divergence between $p$ and $\hat{p}$ is taken as the cost function for training. The choices with top-$N$ scores are selected for a question with $N$ answers.

## 4. EXPERIMENTS

### 4.1. Experimental Setup

The TOEFL listening comprehension test dataset included 963 problems in total, with a train/dev/test split of 717/124/122. There were two versions of each story, manual and ASR transcriptions. The latter was obtained using CMU Sphinx recognizer [23] with a word error rate of 34.32%. The size of the hidden layer for Tree-LSTM [20] and the embedding size of the memory module were both 75. AdaGrad [24] was used with an initial learning rate of 0.002.

### 4.2. Baselines

We compared the proposed model with several baselines as summarized below. The first two were trivial baselines, while the rest were neural network based approaches. In the trivial baselines, we used pre-trained GloVe [25] vectors to obtain the vector representation for each word. Hence, each utterance in the stories, questions and choices could be represented as a fixed-length vector by averaging the word vectors. Cosine distance between vector representations was used to evaluate the similarity between two sentences.

(a) **Question/choice similarity** [1]: With the vector representations for the choices/question mentioned above, the choice most similar to the question was selected.

(b) **Sliding window of utterances** [12, 26]: We slid a window of 5 utterances over the story, and chose the window the most similar to the question as the related information in the story. The choice with the highest similarity to this window was then selected.

(c) **Deep LSTM Reader (DLR)** [16]: We fed the story followed by the question into a Deep LSTM encoder to obtain the representation of each story/question pair. The choices were encoded with the LSTM encoder, and the one whose vector representation with the highest similarity to the story/question pair was selected. Bidirectional LSTM network with hidden layer size of 75 was used as the encoder. All weights in the LSTM units were shared.

(d) **End-to-end Memory Network (MemN2N)** [18]: We slightly modified the original MemN2N to adapt to this task. An embedding matrix was used to embed each choice. We then evaluated the similarity between the output of the last hop and the choice embeddings, and chose the highest one as the answer. The embedding size of the memory network was 300. The hop size was tuned from 1 to 3 with the development set.

(e) **Attention-based Multi-hop Recurrent Neural Network (AMRNN)** [1]: This is the approach used by the previous work. 1-of-N encoding for each word in the question and the choices were entered to a bidirectional GRU network to obtain the vector representations for the question, $V_Q$, and the choices. The vector representation for each word in the story was entered to a bidirectional GRU network to obtain semantic embedding of each word. The cosine similarity between each of them and $V_Q$ was taken as the attention weight. Using these weights, the vector $V_S$ representing the story/question pair using word/sentence-level attention was computed. Finally, the choice whose vector had the highest similarity score to $V_S$ is selected. The hidden layer size for both the forward and backward GRU networks were 128. The number of hop was tuned from 1 to 3 with the development set.

(f) **Tree-LSTM** [20]: Similar to the proposed HAM but without attention. We fed the story/question, one sentence at a time, into a Tree-LSTM encoder as described in Subsection 3.1, then summed the vectors for the sentences in both the story and the query to obtain the vector representation of each pair. The choices were encoded similarly. The selection was based on the highest similarity to the story/question pair. The size of hidden layer for Tree-LSTM was 50.

### 4.3. Results

We used the accuracy (percentage of questions answered correctly) as our evaluation metric. The models were trained on manual transcriptions of the stories and questions/answers of the training set and tested on the manual (column labeled "Manual") and ASR transcriptions (column labeled "ASR") of the testing set. The results are in Table 1. The upper section (a) and (b) are trivial baselines; the middle section (c)-(f) are neural network based baselines; while the lower section (g1)-(g4) are for the proposed HAM with different hops and attention levels. Due to the non-negligible performance variations for all the neural network based models due to random initialization, for fair comparison, we reported the mean accuracies and standard deviations (in parentheses) over 10 runs.

The relatively low accuracies for the trivial baselines in rows (a) and (b) indicated that it is hard to answer correctly without understanding the story, and the information within a short window was inadequate.

The Deep LSTM Reader in row (c), successful in certain QA tasks, was also low (32.4% and 34.3%), similar to the sliding window in row (b). So simply encoding the story and question with a LSTM encoder was not adequate.

| Model | | Manual | ASR |
|---|---|---|---|
| (a) Question choices | | 24.6 | 24.6 |
| (b) Sliding window | | 33.6 | 31.2 |
| (c) DLR | | 32.4(2.2) | 34.3(3.4) |
| (d) MemN2N | | 45.2(1.9) | 44.4(1.3) |
| (e) AMRNN | (e1) word | 42.5(5.0) | 40.1(4.5) |
| | (e2) sentence | 42.4(3.2) | 42.2(4.1) |
| (f) Tree-LSTM | | 46.5(2.8) | 44.9(2.2) |
| (g) proposed HAM | (g1) phrase(1 hop) | 47.7(2.9) | 47.4(3.4) |
| | (g2) phrase(2 hop) | 49.0(3.3) | **48.8(3.3)** |
| | (g3) sent(1 hop) | **49.1(3.1)** | 48.6(2.8) |
| | (g4) sent(2 hop) | 48.2(2.1) | 47.5(2.3) |

**Table 1**: Test results: rows (a)(b) are trivial baselines, rows (c)-(f) are neural network based baselines, rows (g1)-(g4) are the proposed HAM with different hops and attention levels. Mean accuracies and standard deviations (in parentheses) over 10 runs with random initialization are reported for manual and ASR transcriptions.

The MemN2N in row (d), however, was much better (45.2% and 44.4%). This verified the attention over the stories actually extracted better representations for the stories.

The AMRNN in row (e) combined the attention mechanism with the long distance information accumulation in recurrent networks offered reasonably good results on both word-level (42.5% and 40.1%) and sentence-level (42.4% and 42.2%) attention, but slightly lower than MemN2N in row (d). This implied that sequential representations for stories were inadequate even with attention mechanism.[3]

The Tree-LSTM in row (f) exploiting the hidden structures in the sentences but without attention performed relatively well (46.5% and 44.9%), better than MemN2N, although possibly confused by the irrelevant sentences in the stories. This revealed that encoding the hierarchical structures of the sentences was helpful in understanding them.

The approach HAM proposed in this paper is in part (g), respectively for phrase/sentence-level attention with 1 and 2 hops, all much higher than all baselines above. The 1-hop sentence-level attention model achieved the highest mean accuracy of 49.1% on the manual transcriptions, significantly higher than all baselines, while the 2-hop phrase-level attention model achieved the highest mean accuracy of 48.8% on ASR results, only slightly lower than the former. We also observed that increased hops improved the performance for phrase-level attention, but not for sentence-level attention. This is probably because for phrase-level reasoning, the model first selected the key phrases in the first hop and then changed its attention based on these key-phrases on the second hop. For sentence-level reasoning, only a few key sentences were selected in the first hop, while more hops were not able to find additional key sentences.

---
[3]The previous work [1] showed that AMRNN outperformed MemN2N in which 10 models with random initialization were trained, and the best on the development set was used for testing.

**Phrases for nodes with top 3 attention weights (hop1)**

1. incoming energy `sunlight` that is reflected off that surface back to space
   $\alpha_t = 0.002067$
2. there is not much `reflection` go at all
   $\alpha_t = 0.002067$
3. they `transmit` incoming solar energy down to earth
   $\alpha_t = 0.002060$

**Phrases for nodes with top 3 attention weights (hop2)**

1. the `amount` of solar radiation energy from the sun absorbed by earth and the amount reflected back into space
   $\alpha_t = 0.002221$
2. increasing `area` of low thick cloud the type that reflects a large portion of solar energy back to space and cool the earth
   $\alpha_t = 0.002176$
3. `process` that could control the type of cloud
   $\alpha_t = 0.002163$

**Fig. 5**: The phrases in the story for the nodes with top 3 phrase-level attention weights. The head word of the phrase is colored. The question is "*What is the radiation budget?*", and the correct choice is "*The balance between incoming and reflecting solar energy*".

**Word chunks for nodes with top 3 attentions (hop1)**

1. `cloud` in general also has high albedo
   $\alpha_t = 0.077468$
2. one way we keep track of the radiation budget is `looking` at the albedo of the different surface
   $\alpha_t = 0.074881$
3. `cloud` has a high albedo
   $\alpha_t = 0.070614$

**Word chunks for nodes with top 3 attentions (hop2)**

1. `cloud` in general also has high albedo
   $\alpha_t = 0.085345$
2. one way we keep track of the radiation budget is `looking` at the albedo of the different surface
   $\alpha_t = 0.081568$
3. surface's albedo is the `percentage` of incoming energy that is reflected off that surface back to space
   $\alpha_t = 0.072305$

**Fig. 6**: Same as Fig. 5 except for sentence level attention.

### 4.4. Analysis

Fig. 5 shows an example for the phrases (manual transcription) in the story for the nodes with top 3 phrase-level attentions for hop 1 and 2, respectively. The story was about how the climate system strikes a balance between cooling and heating. The question is *"What is the radiation budget?"*, and the correct choice is *"The balance between incoming and reflecting solar energy"*. We can see how the model selected in hop 1 the key phrases related to the definition of *radiation budget*, such as *sunlight, transmission* and *reflection*. In hop 2, some other longer phrases including the abstract definition of *radiation budget* were selected. Fig. 6 is the same except for word chunks for nodes with top 3 sentence-level attentions. We see the selected sentences (word chunks) didn't change much in the hop 2, but the weights for the selected sentences became higher, which may have incorrectly selected sentences and further emphasized them. This may explain why for sentence-level attention hop 2 gave a slightly lower accuracy.

We further analyzed the performance of several models for different types of questions. The TOEFL questions can be divided into 3 categories [27, p.123-p.153]. Type1 questions are for basic comprehension of the story. Type2 questions go beyond basic comprehension, but test the understanding of the functions of utterances or the attitude the speaker expresses. Type3 questions further require the ability of making connections between different parts of the story, making inferences, drawing conclusions, or forming generalizations. We labeled the question types manually on the dataset, resulting in 81 Type1 questions, 15 Type2 questions, and 26 Type3 questions. The results for the different models for the different types of questions are in Fig. 7. We see Type3 questions received the highest scores across all models, suggesting that these neural network based models performed better on relatively difficult reasoning questions than more factoid questions. Also, the proposed HAM outperformed other baselines on Types 1 and 3 questions, verifying that hierarchical attention is useful. On the other hand, the proposed HAM has the lowest accuracies on Type2 questions, probably because they are the most difficult for neural network based models.

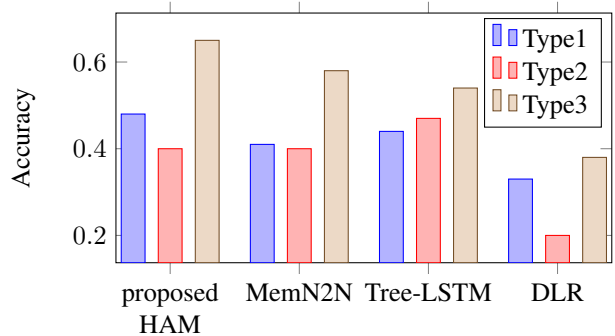

**Fig. 7**: Performance of different neural network based models for different types of questions. 2-hop phrase-level attention was used for the proposed HAM.

### 5. CONCLUSIONS

In this work we presented a Hierarchical Attention Model (HAM) over tree-structured sentence representations, and showed it offered improved performance for machine comprehension of spoken content, based on the TOEFL listening comprehension task. Compared to other neural network based models, the proposed model utilizes multi-hopped attention over tree-structured rather than sequential representations, so information of multiple granularity can be better extracted. This approach is robust with respect to ASR errors, since performance with $34.32\%$ of ASR word error rate was found to be almost the same as those with reference transcriptions.

### 6. ACKNOWLEDGEMENTS

We thank Bo-Hsiang Tseng for assistance with providing us the TOEFL listening comprehension dataset and the detailed statistics of AMRNN [1], and Juei-Yu Chang, An Huang, Yu-An Chen and Ping-Hsuan Tsai for labeling the 3 different types of questions in the testing set.

# 7. REFERENCES

[1] B.-H. Tseng, S.-S. Shen, H.-Y. Lee, and L.-S. Lee, "Towards machine learning comprehension of spoken content: Initial toefl listening comprehension test by machine," in *INTERSPEECH*, 2016.

[2] P. R. C. i Umbert, J. Turmo Borràs, and L. M. Villodre, "Spoken question answering," .

[3] P. R. C. i Umbert, *Factoid question answering for spoken documents*, Ph.D. thesis, Universitat Politècnica de Catalunya, 2012.

[4] S.-R. Shiang, H.-Y. Lee, and L.-S. Lee, "Spoken question answering using tree-structured conditional random fields and two-layer random walk.," in *INTERSPEECH*, 2014, pp. 263–267.

[5] B. Hixon, P. Clark, and H. Hajishirzi, "Learning knowledge graphs for question answering through conversational dialog," in *Proceedings of the the 2015 Conference of the North American Chapter of the Association for Computational Linguistics: Human Language Technologies, Denver, Colorado, USA*, 2015.

[6] P. R. Comas, J. Turmo, and L. Màrquez, "Sibyl, a factoid question-answering system for spoken documents," *ACM Transactions on Information Systems (TOIS)*, vol. 30, no. 3, pp. 19, 2012.

[7] J. Turmo, P. R. Comas, S. Rosset, L. Lamel, N. Moreau, and D. Mostefa, "Overview of qast 2008," in *Workshop of the Cross-Language Evaluation Forum for European Languages*. Springer, 2008, pp. 314–324.

[8] A. Bordes, S. Chopra, and J. Weston, "Question answering with subgraph embeddings," *arXiv preprint arXiv:1406.3676*, 2014.

[9] N. P. Er and I. Cicekli, "A factoid question answering system using answer pattern matching.," in *IJCNLP*, 2013, pp. 854–858.

[10] M. Iyyer, J. L. Boyd-Graber, L. M. B. Claudino, R. Socher, and H. Daumé III, "A neural network for factoid question answering over paragraphs.," in *EMNLP*, 2014, pp. 633–644.

[11] A. Fader, L. Zettlemoyer, and O. Etzioni, "Open question answering over curated and extracted knowledge bases," in *Proceedings of the 20th ACM SIGKDD international conference on Knowledge discovery and data mining*. ACM, 2014, pp. 1156–1165.

[12] M. Tapaswi, Y. Zhu, R. Stiefelhagen, A. Torralba, R. Urtasun, and S. Fidler, "Movieqa: Understanding stories in movies through question-answering," *CoRR*, vol. abs/1512.02902, 2015.

[13] A. Bordes, N. Usunier, S. Chopra, and J. Weston, "Large-scale simple question answering with memory networks," *CoRR*, vol. abs/1506.02075, 2015.

[14] A. Kumar, O. Irsoy, J. Su, J. Bradbury, R. English, B. Pierce, P. Ondruska, I. Gulrajani, and R. Socher, "Ask me anything: Dynamic memory networks for natural language processing," *CoRR*, vol. abs/1506.07285, 2015.

[15] C. Xiong, S. Merity, and R. Socher, "Dynamic memory networks for visual and textual question answering," *CoRR*, vol. abs/1603.01417, 2016.

[16] K. M. Hermann, T. Kociský, E. Grefenstette, L. Espeholt, W. Kay, M. Suleyman, and P. Blunsom, "Teaching machines to read and comprehend," *CoRR*, vol. abs/1506.03340, 2015.

[17] J. Weston, S. Chopra, and A. Bordes, "Memory networks," *CoRR*, vol. abs/1410.3916, 2014.

[18] S. Sukhbaatar, A. Szlam, J. Weston, and R. Fergus, "Weakly supervised memory networks," *CoRR*, vol. abs/1503.08895, 2015.

[19] A. Graves, A.-R. Mohamed, and G. E. Hinton, "Speech recognition with deep recurrent neural networks," *CoRR*, vol. abs/1303.5778, 2013.

[20] K. S. Tai, R. Socher, and C. D. Manning, "Improved semantic representations from tree-structured long short-term memory networks," *CoRR*, vol. abs/1503.00075, 2015.

[21] D. Chen and C. D. Manning, "A fast and accurate dependency parser using neural networks.," in *EMNLP*, 2014, pp. 740–750.

[22] S. Hochreiter and J. Schmidhuber, "Long short-term memory," *Neural Comput.*, vol. 9, no. 8, pp. 1735–1780, Nov. 1997.

[23] W. Walker, P. Lamere, P. Kwok, B. Raj, R. Singh, E. Gouvea, P. Wolf, and J. Woelfel, "Sphinx-4: A flexible open source framework for speech recognition," Tech. Rep., Mountain View, CA, USA, 2004.

[24] J. Duchi, E. Hazan, and Y. Singer, "Adaptive subgradient methods for online learning and stochastic optimization," *Journal of Machine Learning Research*, vol. 12, no. Jul, pp. 2121–2159, 2011.

[25] J. Pennington, R. Socher, and C. D. Manning, "Glove: Global vectors for word representation.," *EMNLP*, vol. 14, pp. 1532–43, 2014.

[26] M. Datar, A. Gionis, P. Indyk, and R. Motwani, "Maintaining stream statistics over sliding windows," *SIAM J. Comput.*, vol. 31, no. 6, pp. 1794–1813, June 2002.

[27] *The official guide to the TOEFL test / ETS*, McGraw-Hill, New York, NY, 2012.